\documentclass{llncs}

\usepackage[dvips]{graphicx} 
\usepackage{isolatin1}       
\begin{document}

\mainmatter

\title{
Dominance Based Crossover Operator \\ 
for Evolutionary Multi-objective Algorithms\\
{\bf DRAFT VERSION}
}

\vspace{0.5cm}

\author
{ Olga Rudenko, Marc Schoenauer}

\vspace{0.3cm}
\institute{
TAO Team, INRIA Futurs \\
LRI, bat. 490, Université Paris-Sud\\
91405 Orsay Cedex, France\\
Olga.Roudenko@lri.fr, Marc.Schoenauer@inria.fr
}

\maketitle 

\pagestyle{empty}

\begin{abstract}
\noindent

In spite of the recent quick growth of the Evolutionary
Multi-objective Optimization (EMO) research field, there has been few
trials to adapt the general variation operators to the particular
context of the quest for the Pareto-optimal set. The only exceptions
are some mating restrictions that take in account the distance between
the potential mates -- but contradictory conclusions have been reported.
This paper introduces a particular mating restriction for Evolutionary
Multi-objective Algorithms, based on the Pareto dominance relation:
the partner of a non-dominated individual will be preferably chosen
among the individuals of the population that it dominates. Coupled
with the BLX crossover operator, two different ways of generating
offspring are proposed.  This recombination scheme is validated within
the well-known NSGA-II framework on three bi-objective benchmark
problems and one real-world bi-objective constrained optimization
problem. An acceleration of the progress of the population toward the
Pareto set is observed on all problems.

\end{abstract}

\section*{Introduction}
The idea of {\em restricted mating} is not new in Evolutionary
Computation: Goldberg~\cite{Goldberg:89} already suggested to forbid,
or at least restrict, the crossover between too different individuals
(i.e. that are too far apart for some distance on the genotypic
space) -- which makes sense for single-objective problems as soon as
the population has started to accumulate on different fitness peaks,
as recombining individuals from different peaks would often
lead to lethal individuals.
This idea has been transposed in the framework of Evolutionary
Multi-objective Algorithms (EMAs)  by 
Hajela and Lin~\cite{HajelaLin:92}, and by Fonseca and Fleming~\cite{FF93}.
Nevertheless, Zitzler and Thiele~\cite{ZitzlerThiele:PPSN98} did not
observe any improvement when mating similar 
individuals.
On the other hand,  Horn et
al.~\cite{Horn:al:94} present an argument supporting mating of 
dissimilar individuals: in the
multi-objective framework, because the population diversity is
enforced, the information provided by very different
solutions can be combined in such way that a new type of (good)
compromises can hopefully be obtained. 
Nevertheless, Schaffer reported the absence of the
improvement when mating dissimilar individuals.  
To sum up, no clear conclusion can be drawn from existing experiments
on the usefulness of restricted mating based on the (dis)similarity
between mates. 

On a real-world design problem, using a very specific 
representation, Wildman et Parks~\cite{WildmanParks:EMO2} have
investigated different pairing strategies based on maximizing or
minimizing different similarity measures. In particular, the
similarity in the sense of the dominance rank has been considered, and
enforcing the mating of the individuals from the elite archive with 
the individuals from the population, in an archive-based EMA, has been
observed to be beneficial.

However, in all studies enumerated above, the efficiency of the
proposed mating restrictions has been measured by the
quality of the final non-dominated solutions, without addressing the
issue of computational time.  In this paper, we propose a restricted
mating strategy whose main effect is to accelerate the progress of the
population of an EMA toward the Pareto set. The main idea is fairly
simple, and 
consists in using the Pareto dominance relation when choosing a mate
for the best (non-dominated) individuals. Note that a more detailed
presentation (in French) can be found in \cite{these-Olga}.

The paper is organized as follows.  Next section briefly
introduces evolutionary multi-objective optimization, and describes
in more detail the NSGA-II algorithm, one of the best performing EMA
to-date, that will be used in all experiments.  Two slightly
different implementations of the dominance-based crossover operator
are then proposed in Section \ref{xover}, based on BLX-$\alpha$
crossover, used throughout this study.  Section \ref{experiments}
presents some 
experimental results witnessing the acceleration of the progress
toward the Pareto set when using the proposed mating
restrictions. Finally, Section \ref{conclusion} gives some guidelines
for a more rigorous and 
complete validation of the proposed  strategy, as well
as for its possible refinements.

\section{Evolutionary Multi-objective Optimization}
\label{emo}

Multi-objective optimization aims at simultaneously optimizing several
contradictory objectives. For such kind of problems, there does not
exist a single optimal solution, and compromises have to be made.

An element of the search space $x$ is said to {\em Pareto-dominate}
another element $y$ if $x$ is not worse than $y$ with respect to all
objectives, and is strictly better than $y$ with respect to at least
one objective. The set of all elements of the search space that are
not Pareto-dominated by any other element is called the {\em
Pareto set} of the multi-objective problem at hand: it represents the
best possible compromises with respect to the contradictory
objectives.

Solving a multi-objective problem amounts to choose one solution among
those non-dominated solutions, and some decision arguments have to be
given. Unlike classical optimization methods, that generally find one
of the Pareto optimal solutions by making the initial optimization
problem single-objective, EMAs are to-date the only algorithms that
directly search for the whole Pareto set, allowing decision makers
to choose one of the Pareto solutions with more complete information.

\subsection{Pareto-based Evolutionary Algorithms}

In order to find a good approximation of the Pareto set (a uniform and
well spread sampling of the non-dominated solutions, close
to the actual Pareto set of the problem at hand), EMAs have to enforce some
progress toward the Pareto set while, at the same time, preserving
diversity between the non-dominated solutions.

Numerous evolutionary methods have been designed in the past years for
the particular task of searching for the Pareto set (the interested
reader will find a good summary in~\cite{book}). The best performing
among them (NSGA-II~\cite{NSGAII:PPSN2000}, SPEA2~\cite{SPEA2:2001},
PESA~\cite{PESA:PPSN2000}) are directly based on the Pareto dominance
relation, that actually ensures progressing toward the non-dominated
set.

Among the diversity preserving techniques, some were transposed to
EMAs from single-objective EAs (such as sharing, for instance),
while others, like the crowding distance 
described in next subsection, are specific to the multi-objective
framework. 

Another recognized important feature of EMAs is
elitism~\cite{ZDT:2000}, directly related to the notion of the Pareto
dominance in EMAs: the non-dominated individuals can be preserved
either by maintaining an archive (SPEA2 and PESA) or by using a
deterministic replacement procedure (NSGA-II).

\subsection{NSGA-II}
\label{NSGA-II}

\begin{figure}[!h]
\begin{center}
\begin{tabular}{ccc}
\includegraphics[width=3.5truecm]{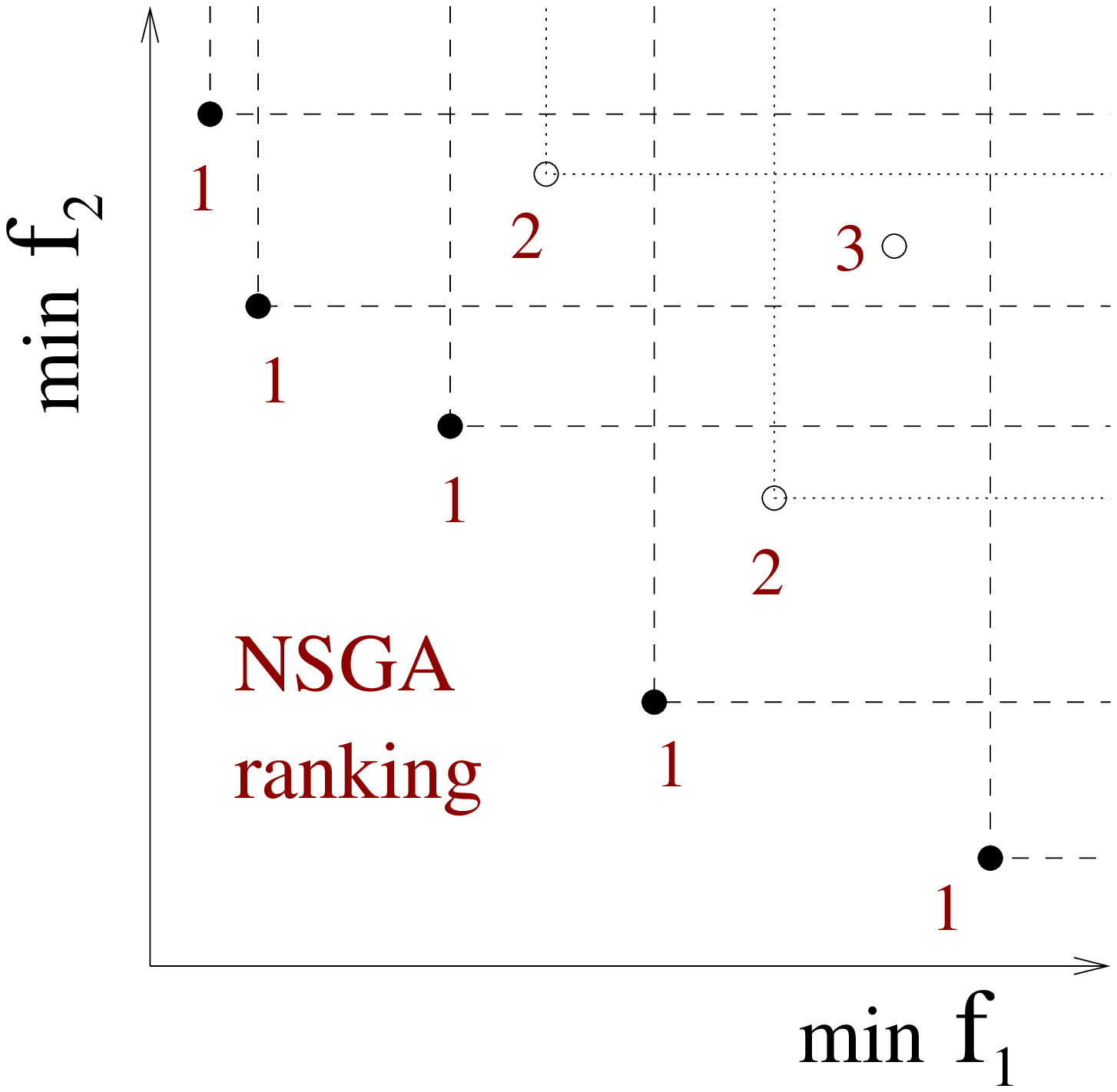}&\mbox{\hspace{0.cm}}&
\includegraphics[width=4.5truecm]{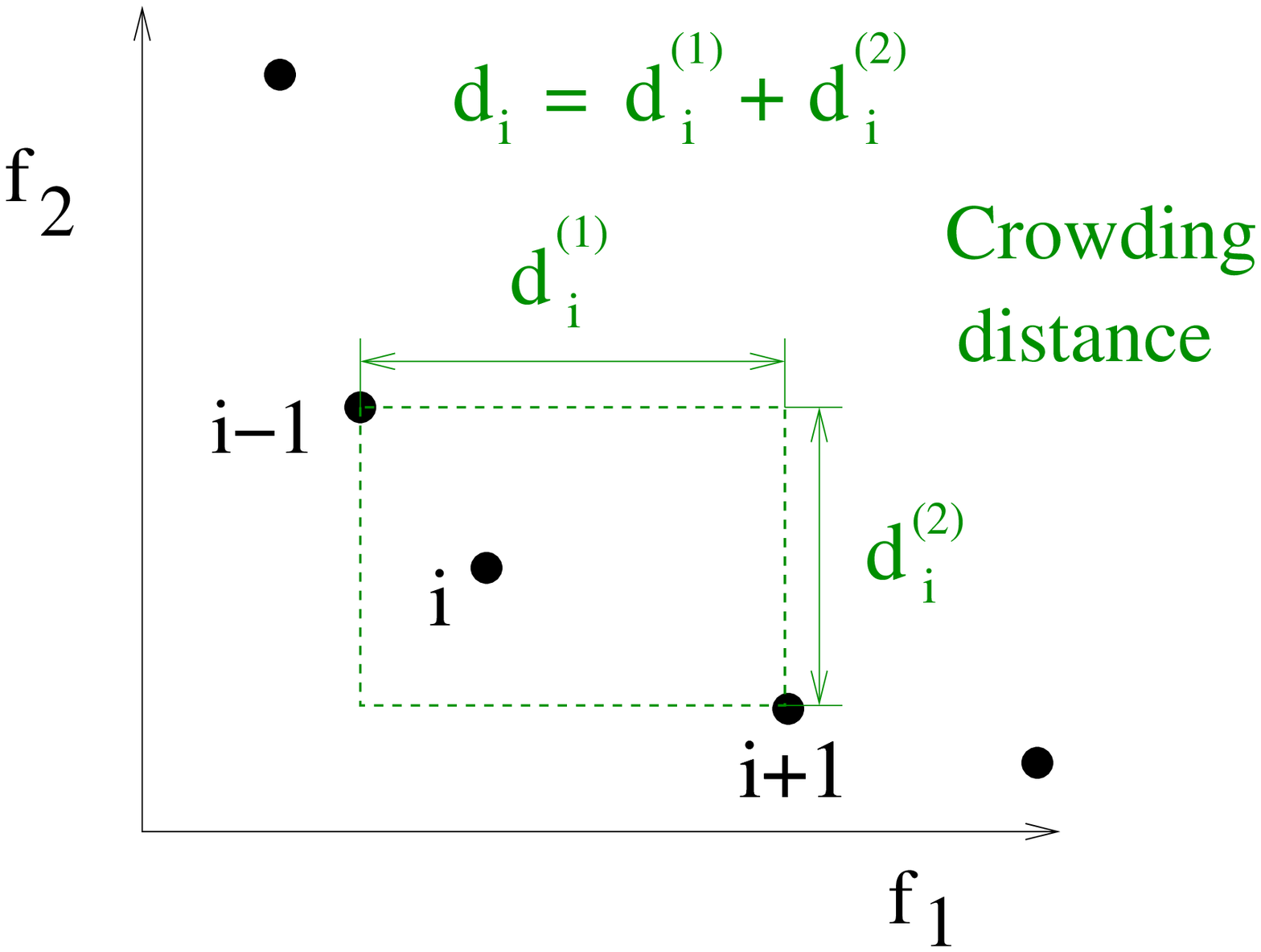}\\
\\
\footnotesize{\bf (a) } Ensuring progress toward the Pareto set & \mbox{\hspace{0.2cm}}& 
\footnotesize{\bf (b) } Preserving diversity technique
\end{tabular}
\end{center}
\vspace{-0.2cm}
\caption{\small\it NSGA-II comparison criteria}
\label{nsga2fig}
\end{figure}

The NSGA-II algorithm has been proposed by Deb et al. in
2001~\cite{NSGAII:PPSN2000}. The progress toward the Pareto set is
here favored by using a selection based on the {\em Pareto ranking},
that divides the population into a hierarchy of non-dominated subsets,
as illustrated by figure~\ref{nsga2fig}(a).  All non-dominated
individuals of the population are first labeled as being of rank~1;
they are then temporarily removed from the population, and the process
is repeated: the non-dominated individuals of the remainder of the
population are given rank~2, and so on, until the whole population is
ranked.

NSGA-II diversity preserving technique is based on the {\it crowding
distance} - one of the possible estimations of the density of the
solutions belonging to the same non-dominated subset. The crowding
distance of each individual $i$ is computed as follows: the
non-dominated subset to which the individual $i$ belongs is ordered
following each of the objectives; for each objective $m$, the distance
$d_i^{(m)}=f_m(i+1)-f_m(i-1)$ between the surrounding neighbors of
individual $i$ according to objective $m$ is computed
(Fig.~\ref{nsga2fig}(b)); the sum over all objectives of these
distances is the crowding distance of individual $i$.

The following comparison operator $\succ$ is then used 
during the Darwinian stages (selection and replacement) of NSGA-II:\\

\noindent
\begin{tabular}{l}
$\quad x \succ  y$ {\bf iff}  $\quad$rank$(x)$ $<$  rank$(y)$\\
\hspace{2cm} {\bf or} $\quad$rank$(x)$ $=$ rank$(y)$\\
\hspace{2.9cm} {\bf and } $\quad$crowding\_dist$(x)$ $>$ crowding\_dist$(y)$$\quad$
\end{tabular}

NSGA-II selection is based on tournament: it chooses
an individual for  reproduction by uniformly drawing $T$ individuals 
(generally, $T=2$) from the population and returning the best of them
with respect to the comparison operator $\succ$. 
NSGA-II replacement is deterministic: 
it consists in merging parents and offspring together,
and choosing the  $N$ best individuals in that global population,
again using comparison operator $\succ$. 
The algorithm NSGA-II is said to be elitist because
the best (less crowded non-dominated)  individuals are preserved from
one generation to another.

\section{Dominance-based crossover}
\label{xover}

The basic mechanism of the proposed mating restriction consists in allowing
the mating of each of the best individuals only with an individual it
dominates (if any), where {\it best individuals} means 
non-dominated individuals when applying NSGA-II, or members of the
archive when using SPEA2 ou PESA.  

The rationale behind using the dominance relation to form the couples for
the crossover is the following.  If $x$  dominates $y$, then $x$ is better
than $y$ for all objectives. Hence, the direction $y \rightarrow x$ is
likely to improve all criteria simultaneously.
Furthermore, a natural continuation of the same idea is to bias the
distribution of the offspring  toward the dominant parent, as better
individuals are more likely to be found close to it.
However, it is clear that success of this idea depends
on the behavior of the objective functions in the region of the
decision space where the mated individuals sit.
 
The resulting crossover, called Dominance-Based Crossover (DBX)  
will proceed as follows: a first mate is chosen using
the usual selection procedure of the EMA at hand (e.g. tournament based
on the $\succ$ operator for NSGA-II). If the chosen individual is non-dominated
and dominates some other individuals in the population, its mate is chosen among
those. Otherwise, the mate is chosen using the usual selection
procedure. In any case, crossover then proceeds with the chosen
operator. 

In this study, the well-known BLX-$\alpha$ crossover ($0<\alpha<1$),
proposed by Eshelman and Schaffer~\cite{Eshelman:Schaffer:93} for 
real representations, has been used.  Formally, given two
parents $(x_i)_{i\in[1,n]}$ and $(y_i)_{i\in[1,n]}$, this operator
produces an offspring by a random linear recombination of  the parents
as follows: 

\vspace{-0.3cm}
\begin{equation}
\label{blx}
\left( (x_i)_{i\in[1,n]}, (y_i)_{i\in[1,n]} \right) \longrightarrow 
\left(\phi_i x_i + (1-\phi_i)  y_i \right)_{i\in[1,n]}, 
\end{equation}
where $\phi_i = U[\alpha, 1+\alpha]$.  In our particular case, given a
non-dominated indi\-vi\-dual $(x_i)_{i\in[1,n]}$ from the NSGA-II
population, two possible strategies will be considered to generate the
offspring:

\vspace{-0.3cm}
\begin{enumerate}

\item {\small\bf Symmetric DBX}: 
The mate $(y_i)_{i\in[1,n]}$ is chosen  
from the list of the individuals dominated by $(x_i)_{i\in[1,n]}$, if
any, by tournament otherwise, and $\phi_i = U[-0.5, 1.5]$ in
Equation~(\ref{blx}).

\item {\small\bf Biased DBX}: 
Similarly, the mate $(y_i)_{i\in[1,n]}$ is chosen  
from the list of the individuals dominated by $(x_i)_{i\in[1,n]}$, if
any, but now $\phi_i = U[0.5, 1.5]$ in Equation~(\ref{blx}), i.e. the
offspring will be closer to the first parent $(x_i)_{i\in[1,n]}$.
\end{enumerate}

\section{Experimental results}
\label{experiments}

\subsection{Experimental conditions}
This section presents some experimental results, on three standard
benchmark problems~\cite{ZDT:2000} and on an
industrial problem~\cite{EAmultipla}. All experiments are run with 
population size 100 and tournament size 2. The two DBX
crossovers  are compared to the
standard BLX-0.5, the crossover rate is set to 0.9. 
The uniform mutation operator is
applied with rate 0.05. The algorithms run for at most
150 (resp. 250) generations for ZDT-benchmarks (resp. 
the industrial problem).

\subsection{Bi-objective ZDT benchmarks}

\begin{figure}[!h]
\vspace{-0.2cm}
\begin{center}
\begin{tabular}{cc}
\includegraphics[width=5truecm]{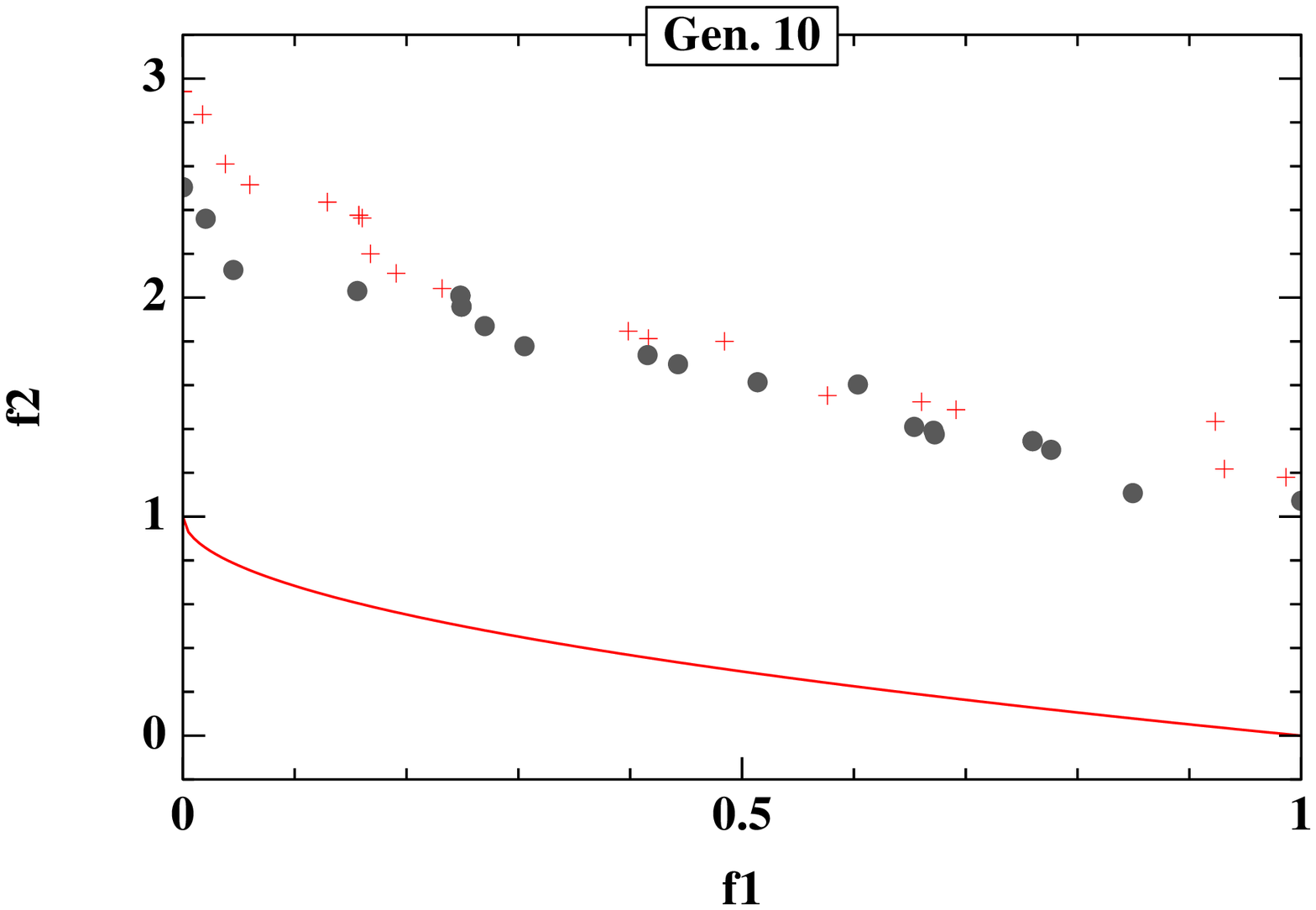}&\includegraphics[width=5truecm]{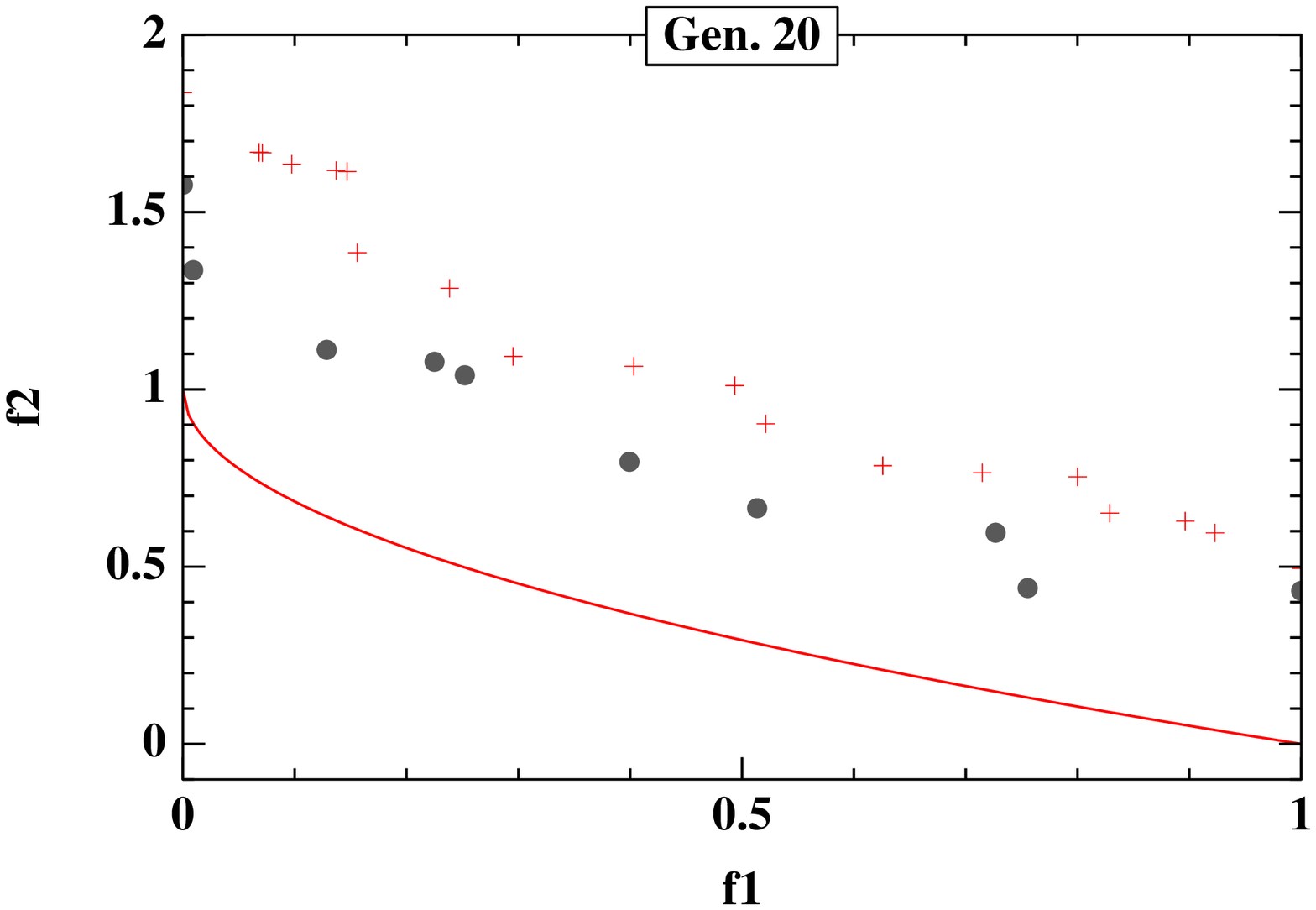}\\
\includegraphics[width=5truecm]{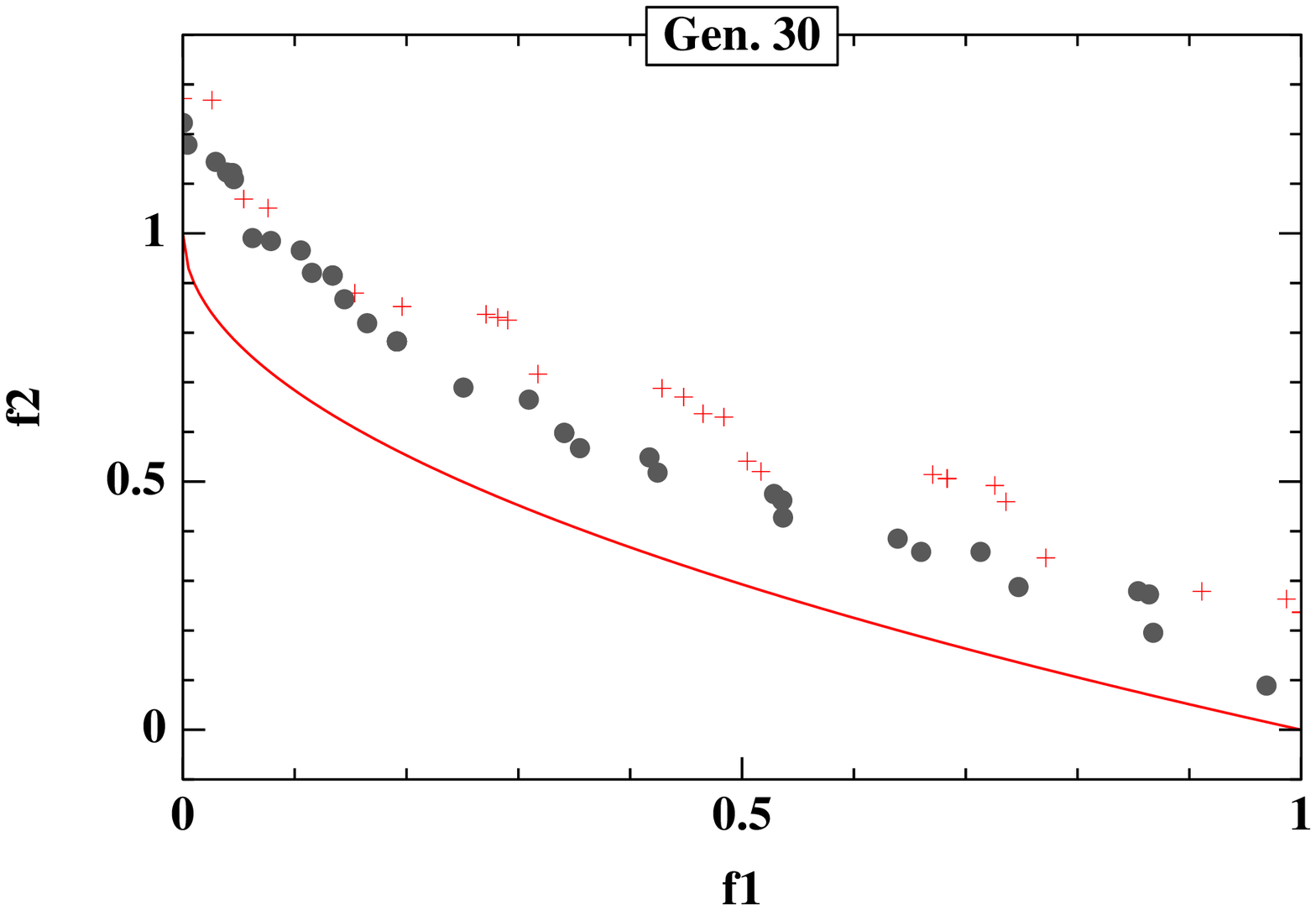}&\includegraphics[width=5truecm]{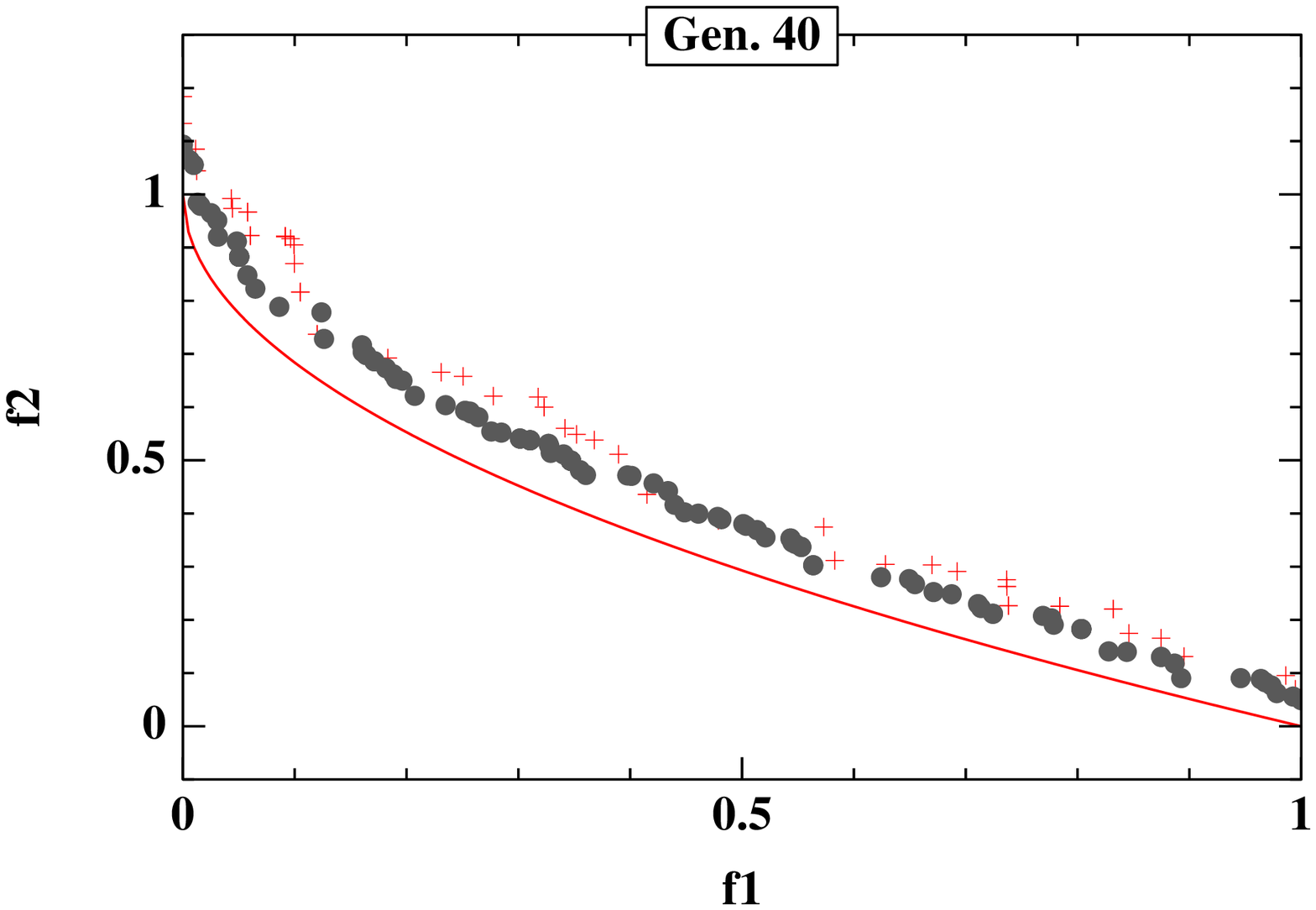}\\
\end{tabular}
\end{center}
\vspace{-0.2cm}
\caption{\footnotesize\it {\bf ZDT1}: the black bullets (for the biased DBX)
and the gray crosses (for the standard BLX-0.5) represent the non-dominated 
individuals over 31 runs at generations 10, 20, 30 and 40.}
\label{zdt1}
\end{figure}

\begin{figure}[!h]
\begin{center}
\begin{tabular}{cc}
\includegraphics[width=5truecm]{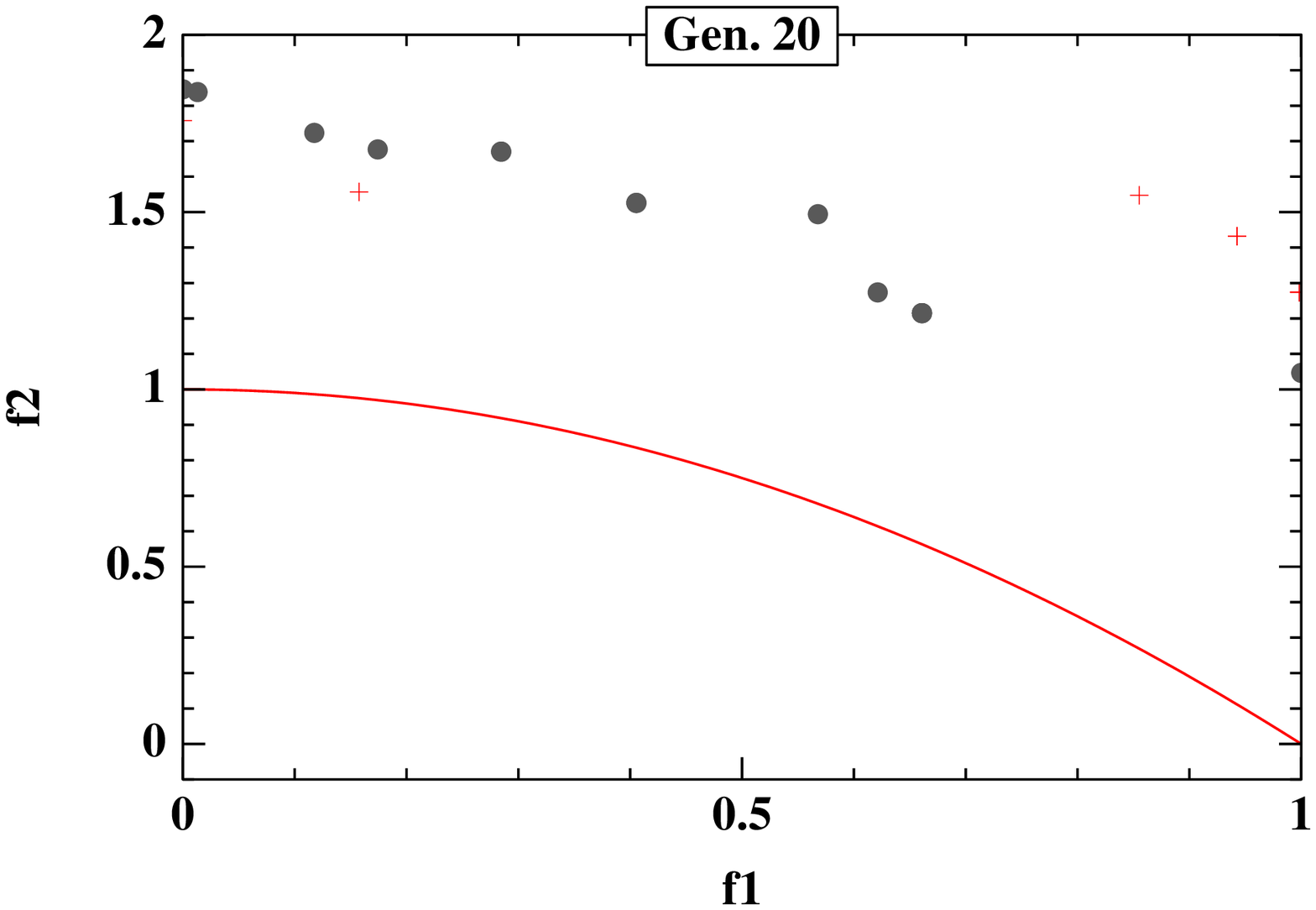}&\includegraphics[width=5truecm]{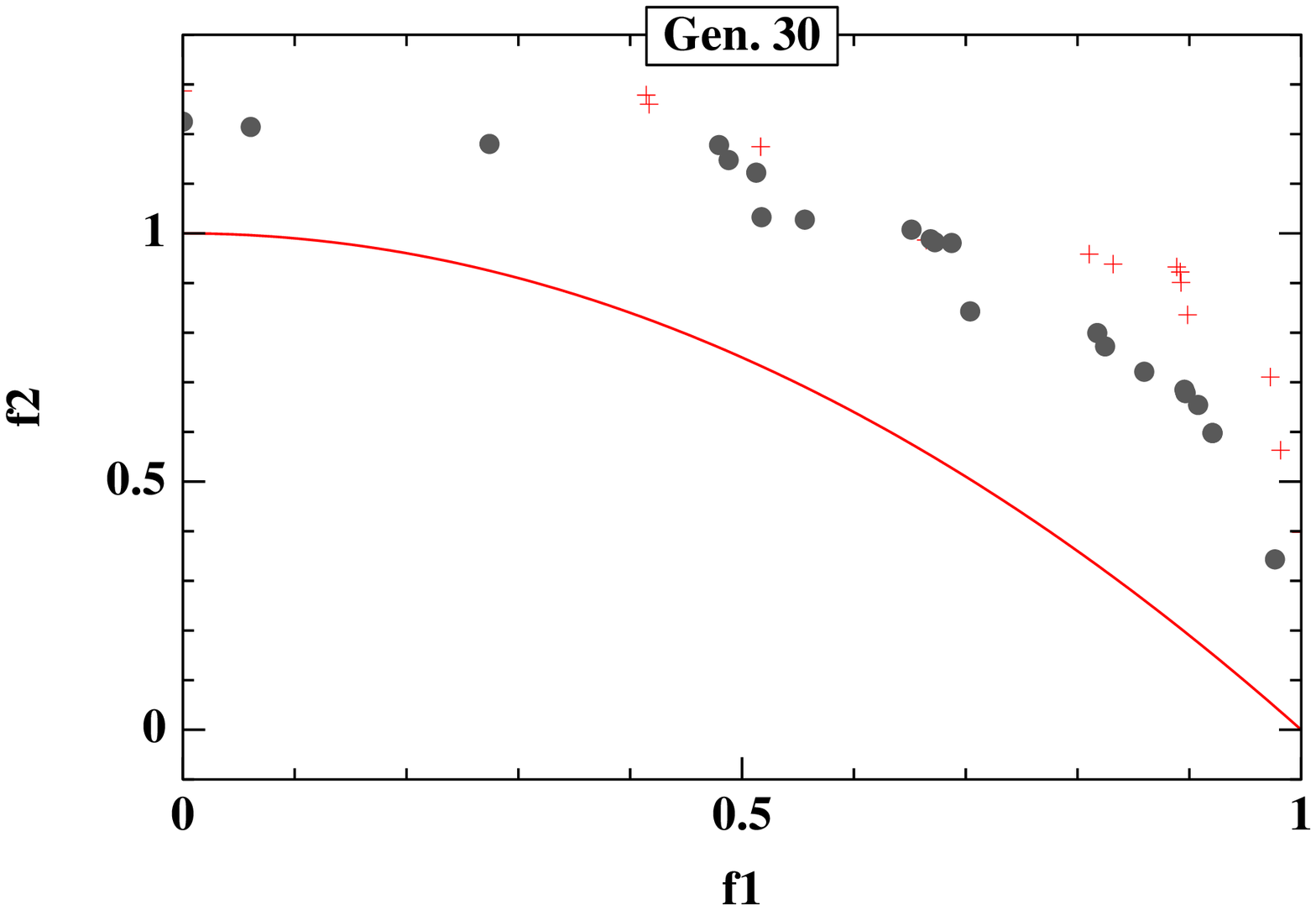}\\
\includegraphics[width=5truecm]{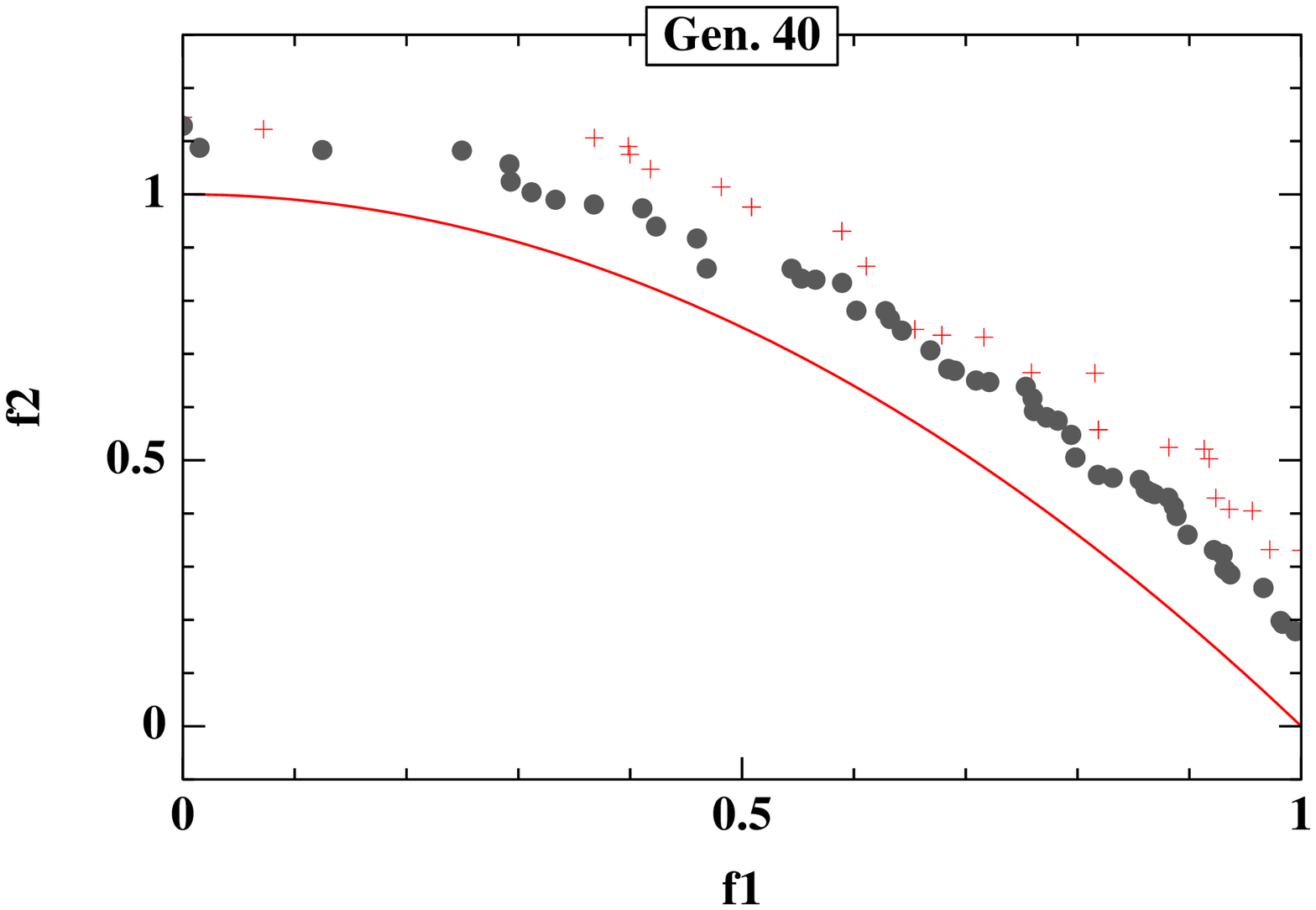}&\includegraphics[width=5truecm]{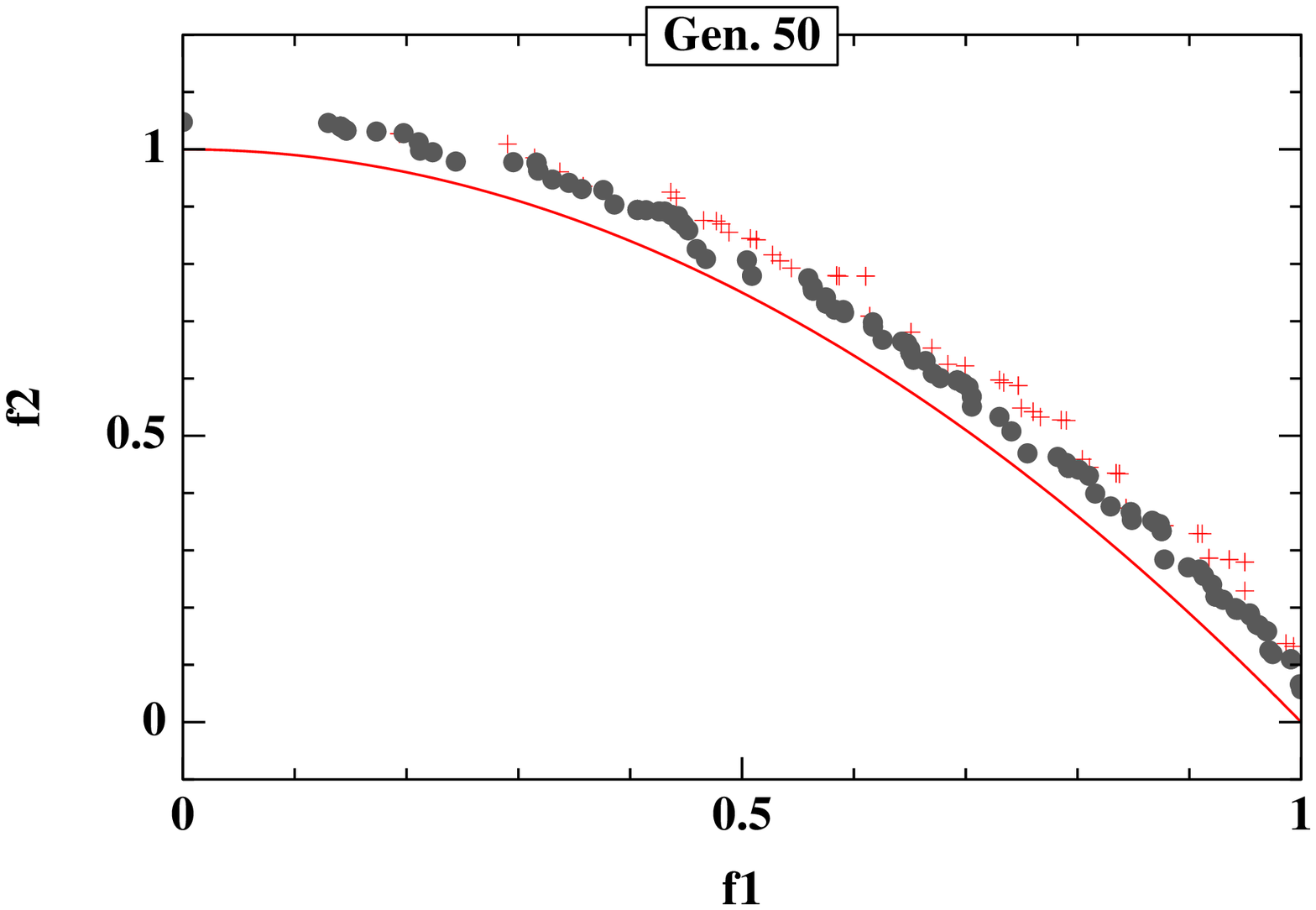}\\
\end{tabular}
\end{center}
\vspace{-0.2cm}
\caption{\small\it {\bf ZDT2}: the black bullets (for the biased DBX)
and the gray crosses (for the standard BLX-0.5) represent the non-dominated 
individuals over 31 runs at generations 20, 30, 40 and 50.}
\label{zdt2}
\end{figure}

\begin{figure}[!h]
\vspace{0.2cm}
\begin{center}
\begin{tabular}{cc}
\includegraphics[width=5truecm]{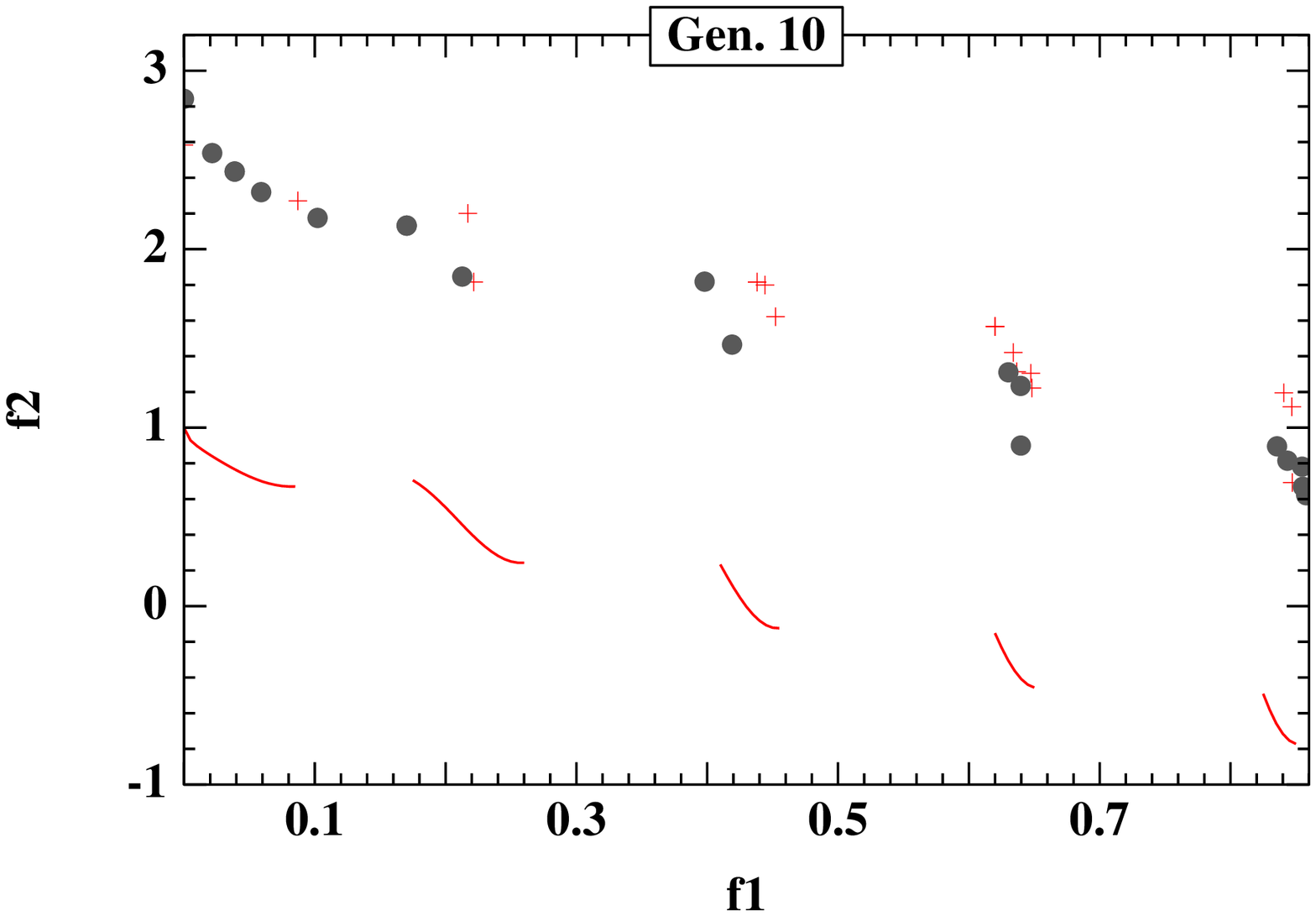}&\includegraphics[width=5truecm]{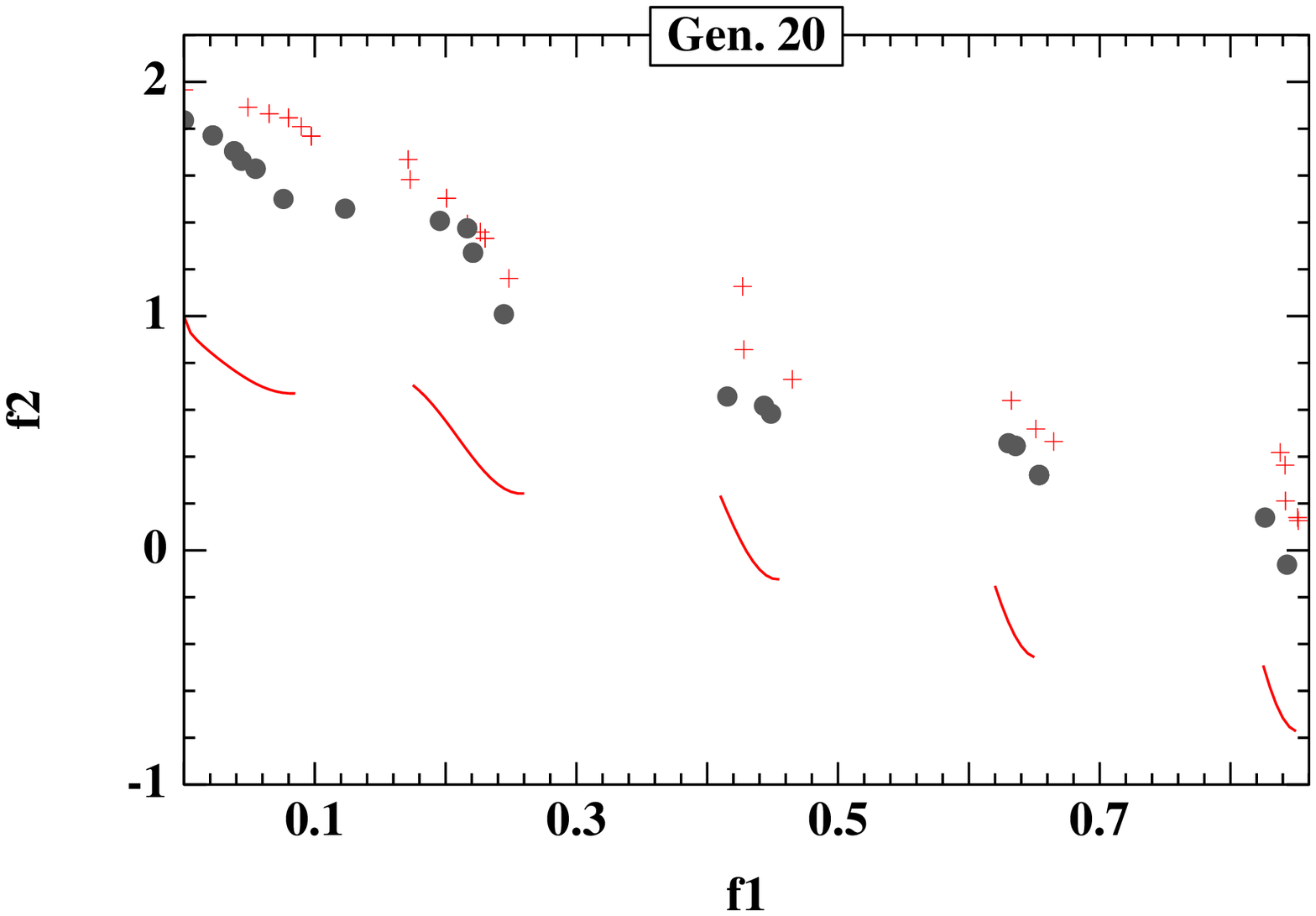}\\
\includegraphics[width=5truecm]{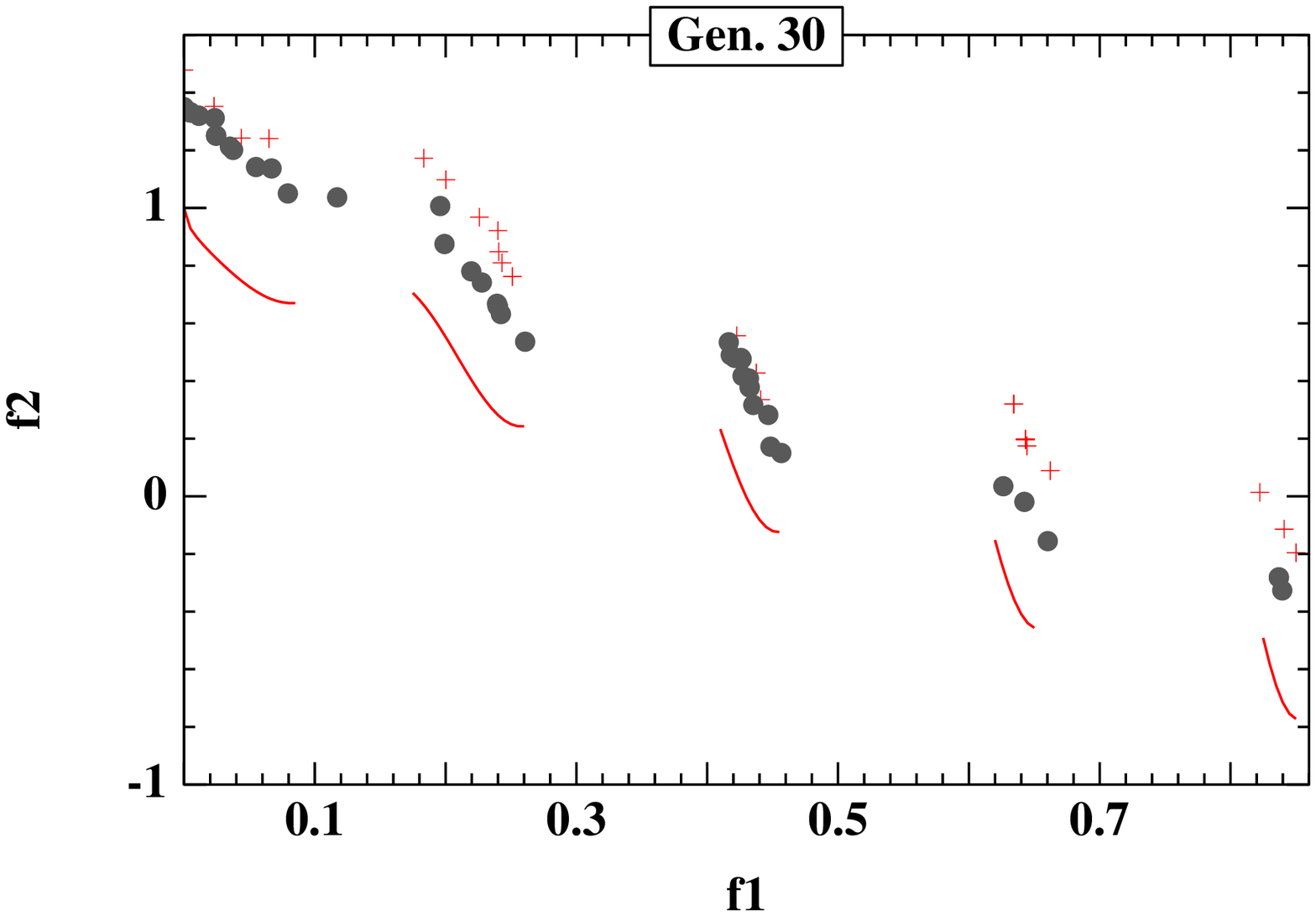}&\includegraphics[width=5truecm]{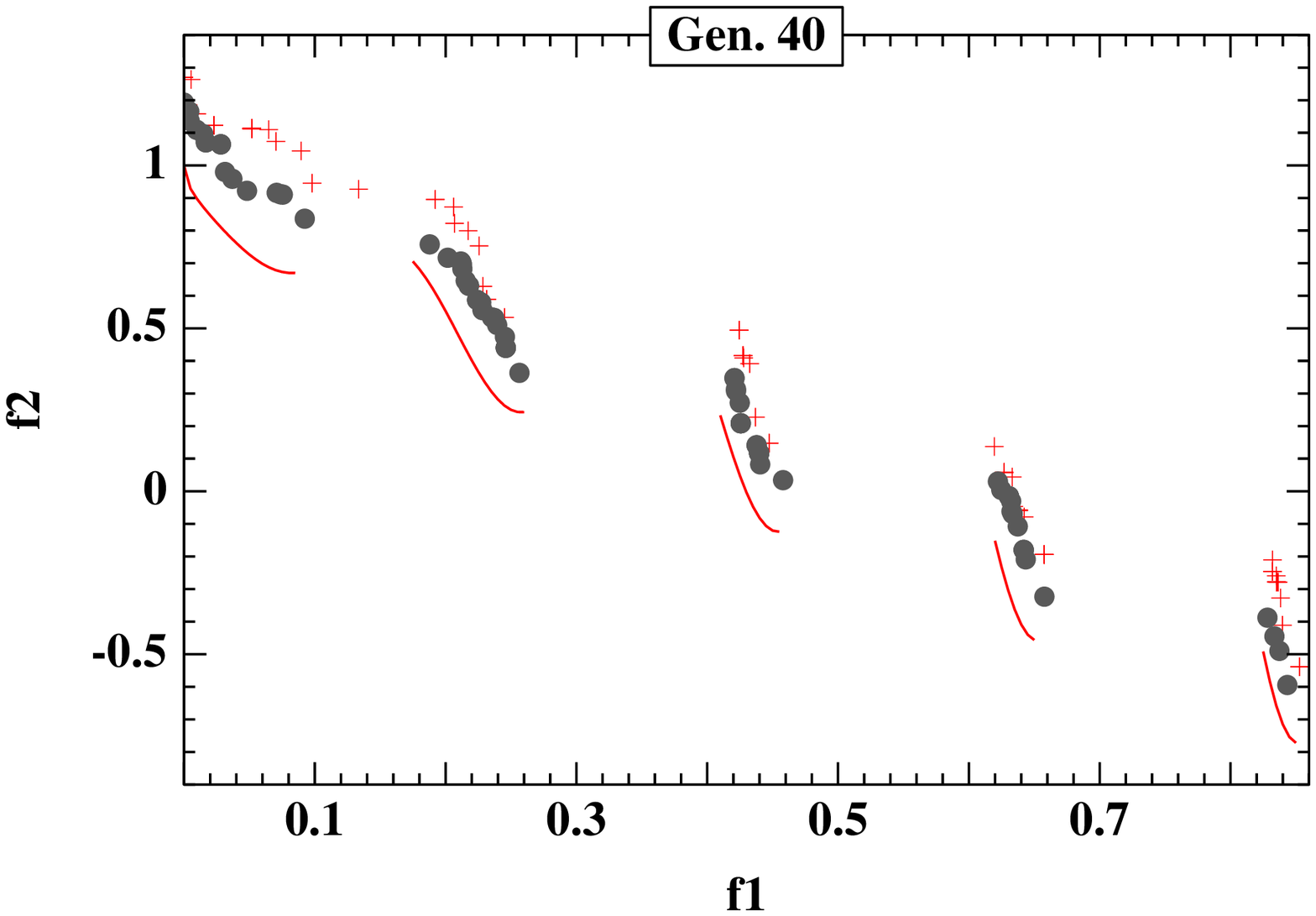}\\
\end{tabular}
\end{center}
\vspace{-0.2cm}
\caption{\small\it {\bf ZDT3}: the black bullets (for the biased DBX)
and the gray crosses (for the standard BLX-0.5) represent the non-dominated 
individuals over 31 runs at generations 10, 20, 30 and 40.}
\label{zdt3}
\end{figure}

For each of ZDT1--ZDT3 test problems 31 NSGAII runs have been performed
for biased DBX, symmetric DBX and standard BLX-0.5 operators 
starting with the same 31 initial populations. 
The non-dominated individuals over all 31 runs have been calculated 
at each 10th generation approximately until the moment when the whole
population is non-dominated, that means that DBX crossover 
is not applied any longer. 
These snapshots corresponding
to biased DBX and standard BLX-0.5 are
shown in the figures~\ref{zdt1}, ~\ref{zdt2} and ~\ref{zdt3}
for ZDT1, ZDT2 and ZDT3 respectively. 
On all three test problems, a small but steady acceleration of the
progress toward the Pareto front is observed when using DBX crossover.
On those problems, very similar results have been obtained with 
the symmetric and biased DBX operators (the snapshots corresponding
to the symmetric DBX are not shown for the space reasons).

\vspace{0.1cm}

\subsection{Constrained bi-objective optimization}

This industrial problem consists in optimizing the structural
geometry, described by 13 continuous parameters, of the front crash
member of the car in order to minimize its mass while maximizing the
internal energy absorbed during a crash (two competitive objectives)
under 8 constraints arising from the acoustic and static mechanical
domains~\cite{EAmultipla}.

The constraints have been handled using the 
so-called {\it infeasibility objective
approach}~\cite{WrightLoosemore:EMO1}:
the aggregated sum of the scaled constraint violations (that can be
viewed as a measure of distance separating each individual from the
feasible region) was considered as an additional optimization
criterion -- the infeasibility objective. 
NSGA-II was hence  applied
to a three-objective optimization problem, with the difference that
the infeasibility objective had a higher priority than both others in
the $\succ$ operator.
In the other words, every individual located  closer to the feasible
region is considered better than any individual further away
from it, regardless the corresponding values of the other objectives, 
the mass and  the absorbed internal energy.

\begin{figure}[!h]
\vspace{-0.2cm}
\begin{center}
\begin{tabular}{cc}
\includegraphics[width=6truecm]{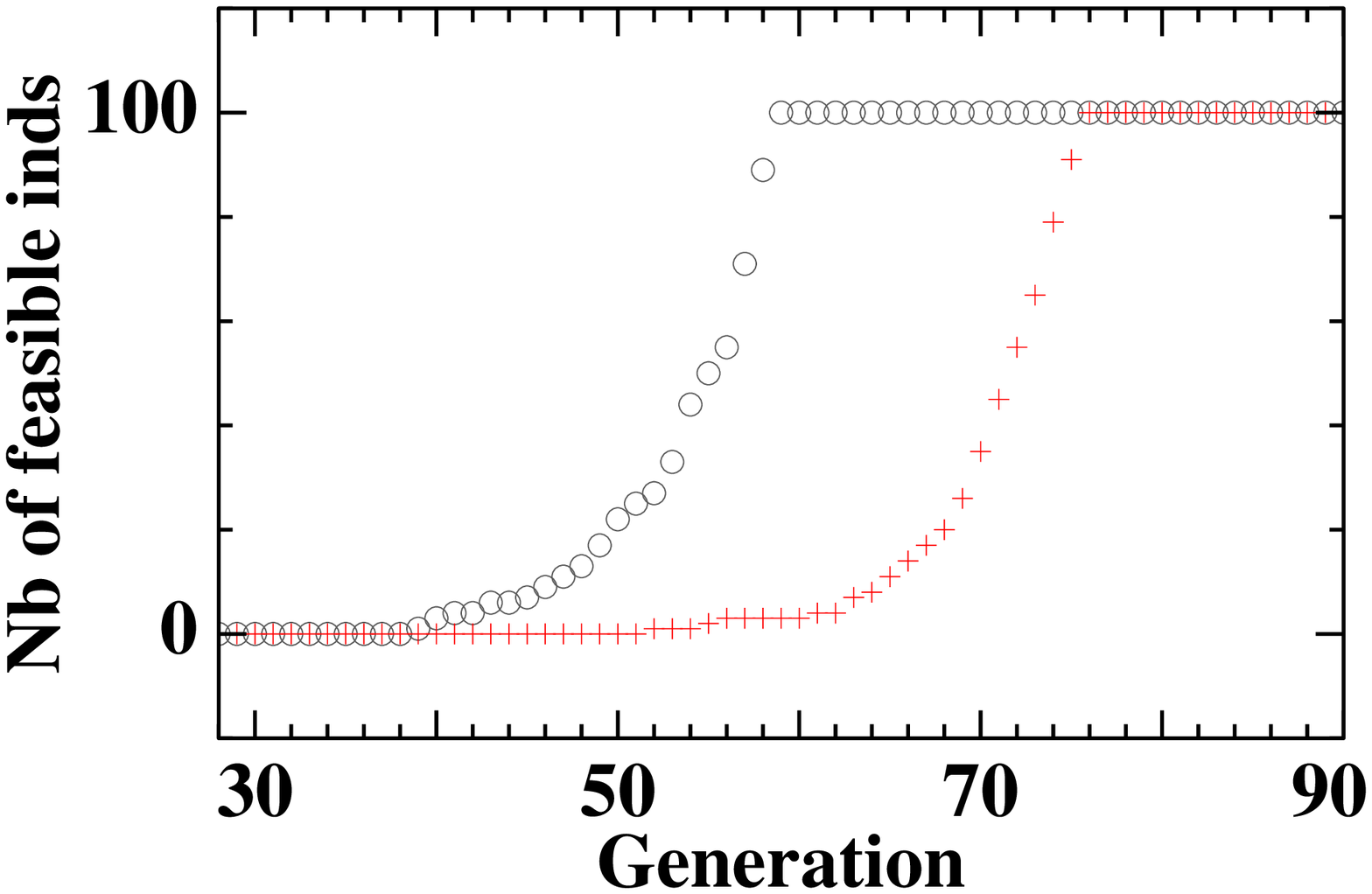}&\includegraphics[width=6truecm]{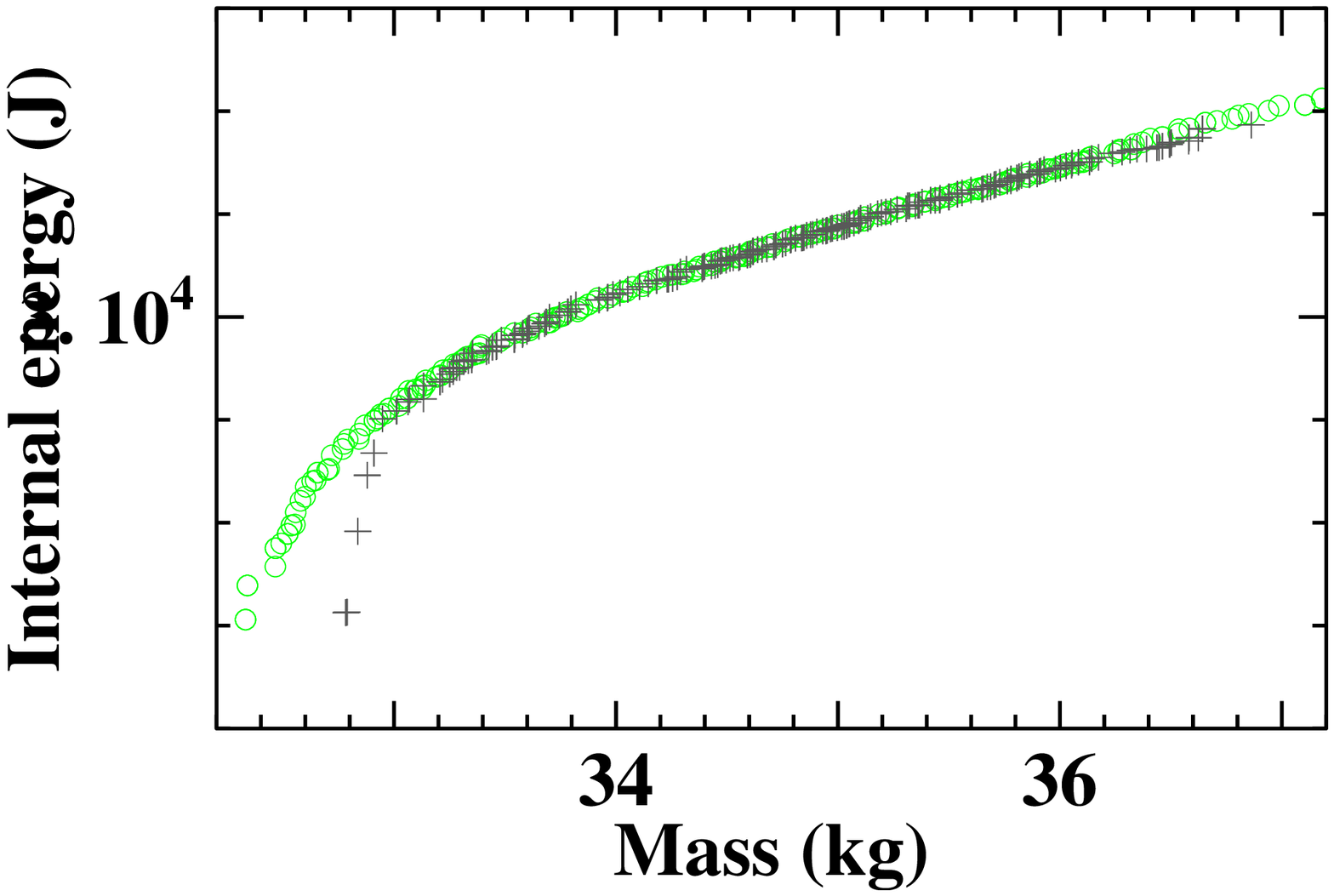}\\
{\small\bf (a)} \tiny Faster reaching the feasible region &
{\small\bf (b)} \tiny Better sampling the Pareto front extremities
\end{tabular}
\end{center}
\vspace{-0.2cm}
\caption{\small\it Effect of the dominance-based mating restriction
  in the presence of constraints:
biased DBX (rings) VS standard BLX-0.5 (crosses)}
\label{constraints}
\end{figure}

DBX operators can be employed in this context exactly like
for unconstrained problems, using the modified $\succ$ comparison.
On every run, the restricted mating allowed faster approach of the
feasible region, as well as faster feasibility of the whole population,
as illustrated by Figure~\ref{constraints}-a. Moreover, some
 differences with the results on the test problems ZDT have
been steadily observed. First, biased DBX was more efficient than
symmetric DBX. Then, and more importantly, the use of mating
restriction not only accelerated the search, it also provided
solutions of better quality at the extremities of the Pareto front,
as it can be seen on figure~\ref{constraints}-b.

\section{Discussion and future work}  
\label{conclusion}

For all four problems considered in this study, the DBX
operators (that only allow the mating of
dominant individuals with individuals they dominate) 
have been shown to accelerate the progress 
of the populations toward the Pareto set. Moreover,
for the optimization of the car front crash  member, it also allowed
finding solutions of better quality at the 
extremities of the Pareto set that could not be reached when using
the usual recombination strategy.

\begin{figure}[!h]
\begin{center}
\begin{tabular}{cc}
\includegraphics[width=5truecm]{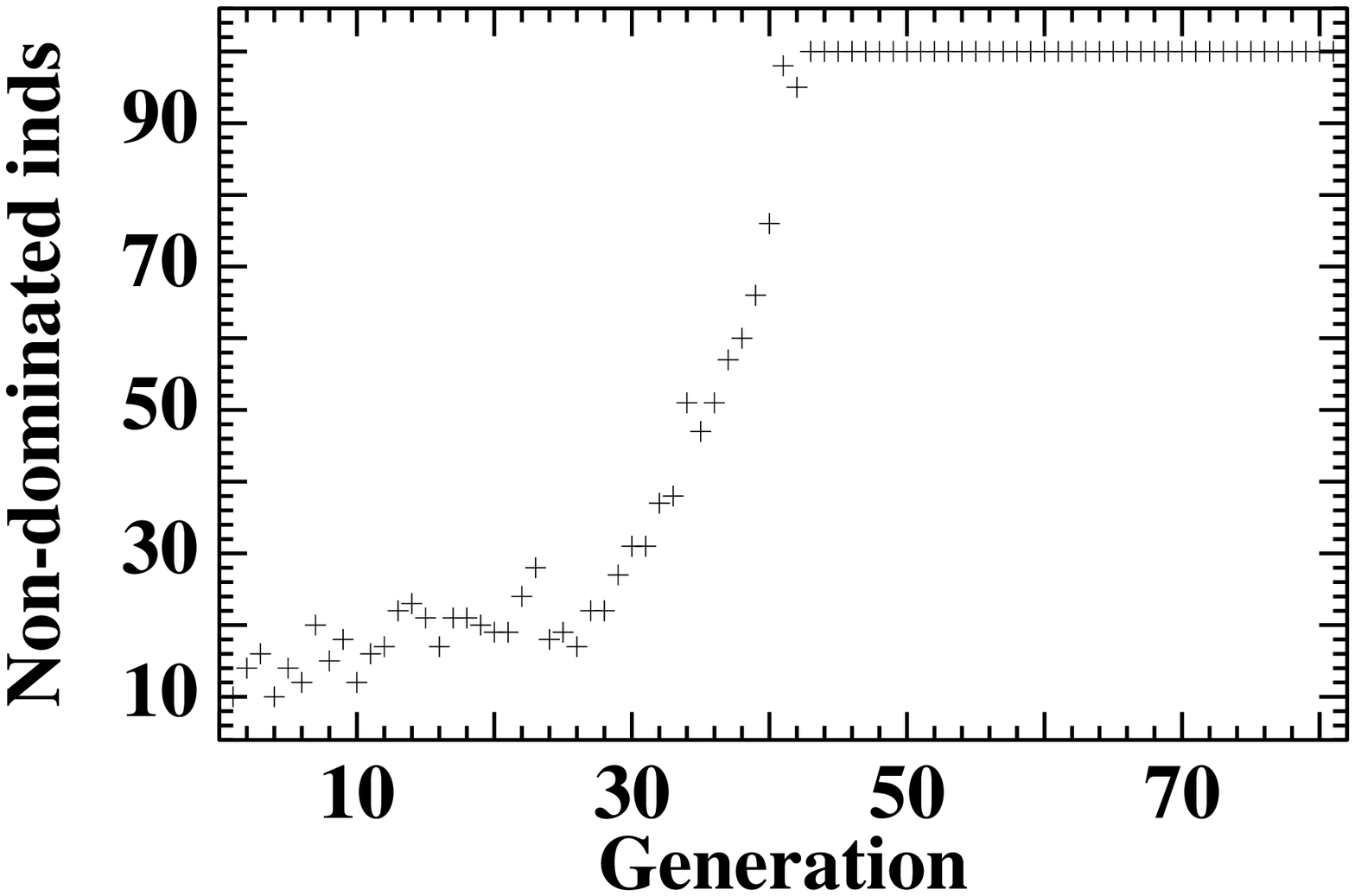}&
\includegraphics[width=5truecm]{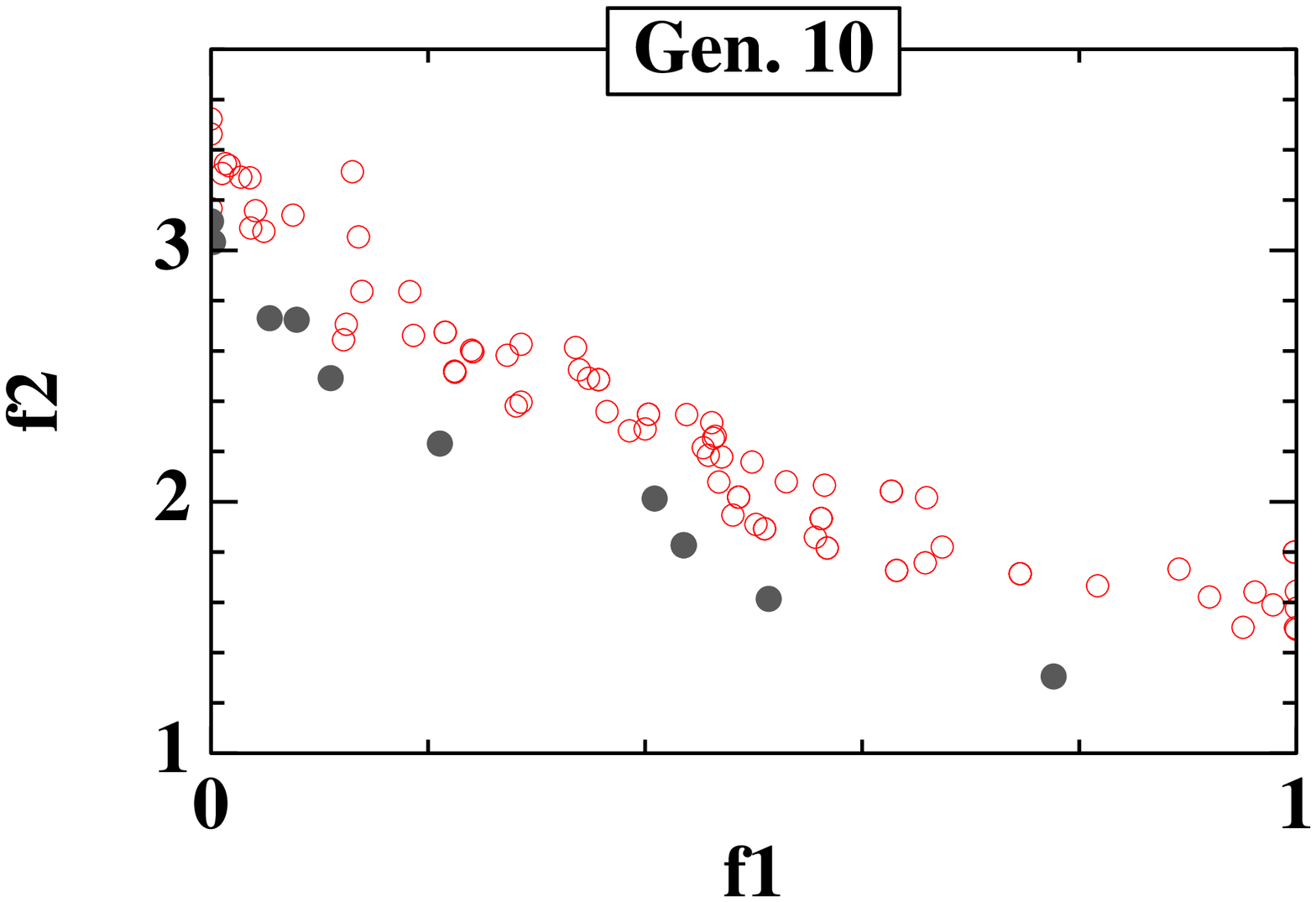}\\
{\bf\small(a)} \tiny Number of non-dominated individuals& 
{\bf\small(b)}  \tiny 12 non-dominated inds at the 10th gen. \\
& \\
\includegraphics[width=5truecm]{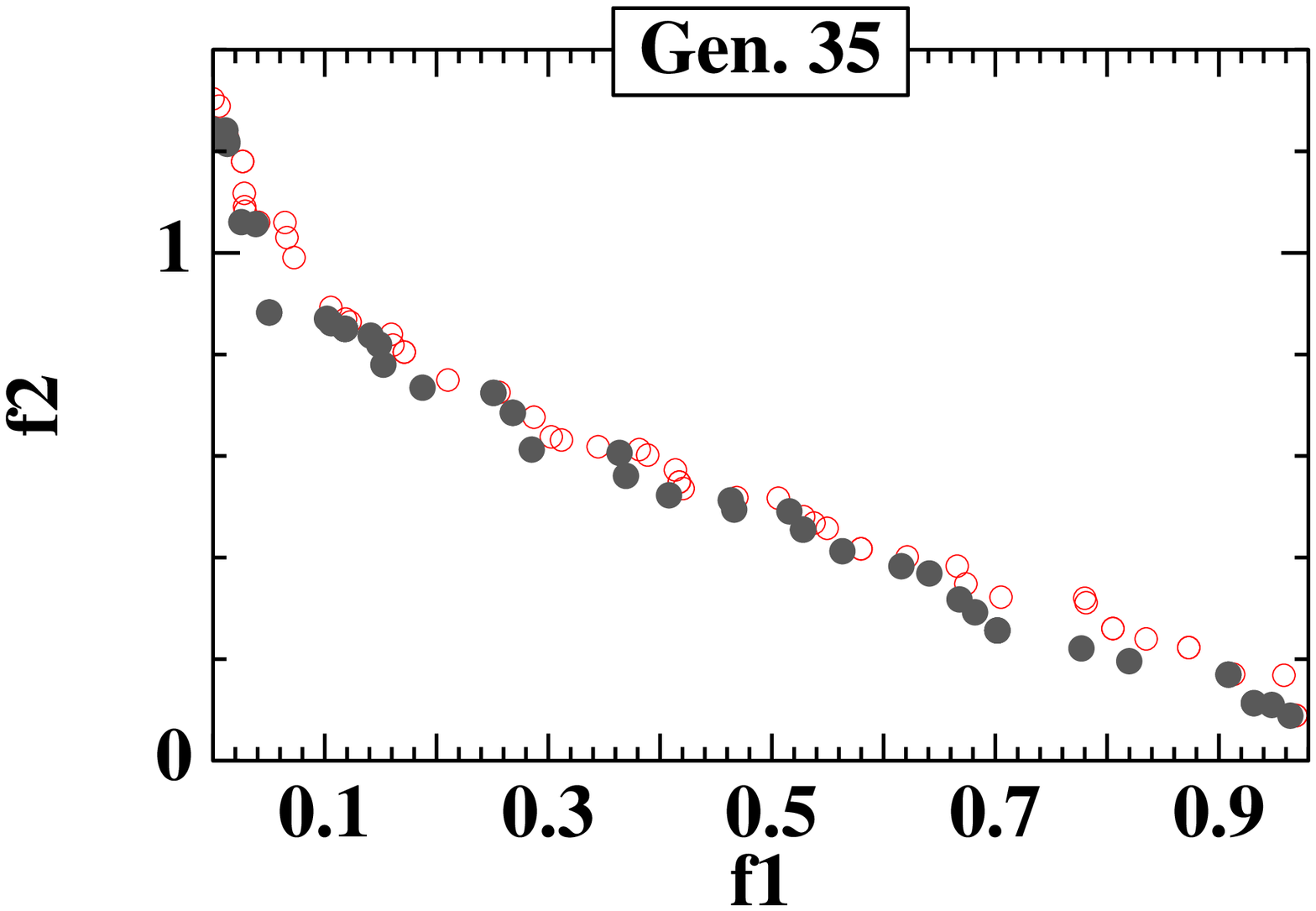}&
\includegraphics[width=5truecm]{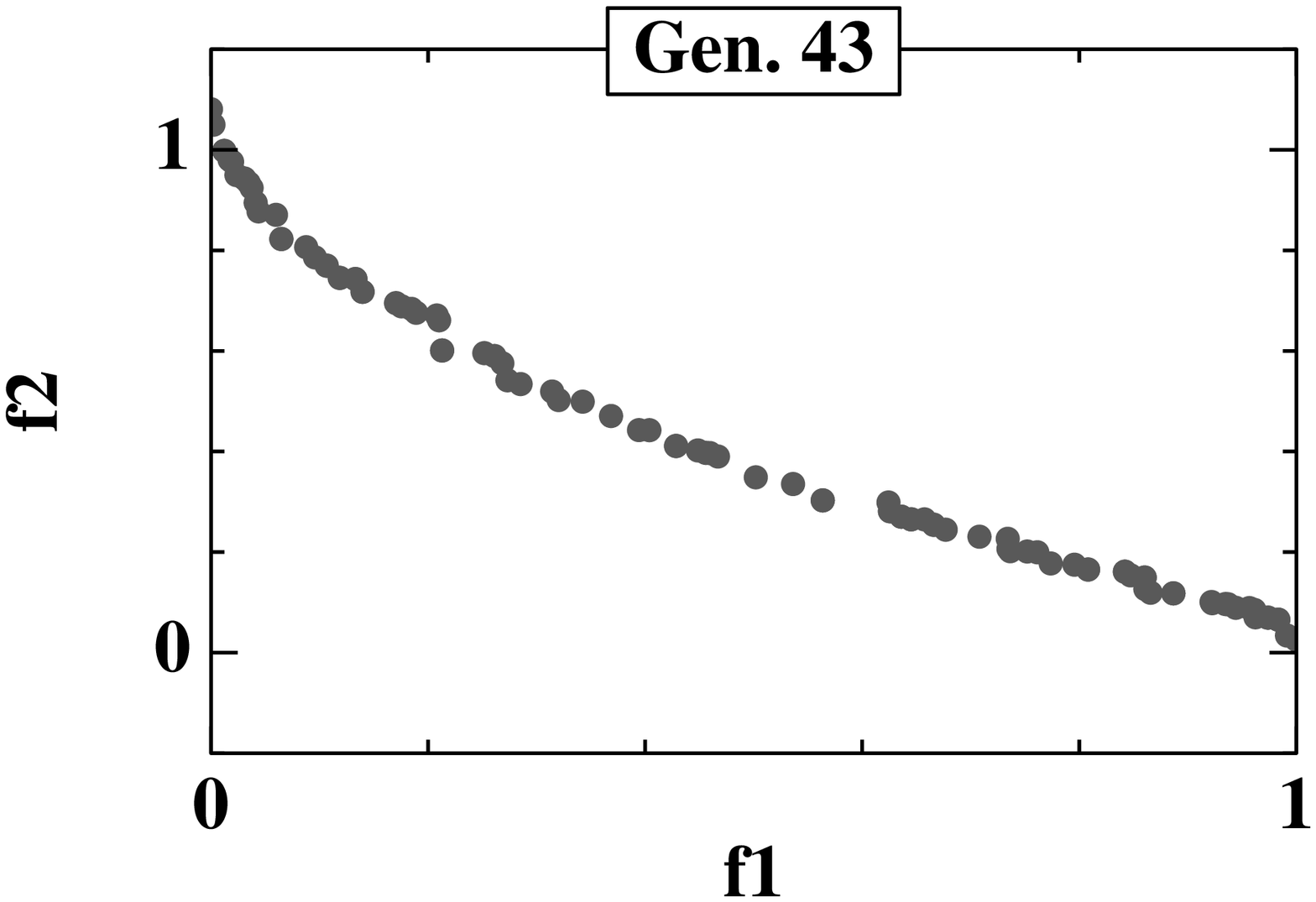}\\
{\bf\small (c)} \tiny  47 non-dominated inds at the 35th gen.& 
{\bf\small (d)} \tiny 100 non-dominated inds at the 43rd gen.
\end{tabular}
\end{center}
\vspace{-0.2cm}
\caption{\footnotesize\it Evolution of the number of non-dominated
  individuals during a run of NSGA-II}
\label{nondNb}
\vspace{-0.5cm}
\end{figure}
The observed acceleration is relatively small, but is
systematic. When observing NSGA-II dynamics (the typical NSGA-II
evolution process is illustrated by Figure~\ref{nondNb}), we realize that
in fact only a small number of DBX crossovers are actually applied.
First, note that restricted mating is applied at most as many
times as there are non-dominated individuals in the population.
There are few non-dominated individuals at the beginning of evolution 
(Fig.~\ref{nondNb}-a,b),  but each of them dominates
a lot of other individuals (Fig.~\ref{nondNb}-b).  
As the population gets closer to the Pareto set, the number of
non-dominated individuals rapidly increases (Fig.~\ref{nondNb}-a,c), 
but each of them dominates only a few individuals if any (Fig.~\ref{nondNb}-c). 
Finally, when the whole 
population is non-dominated (Fig.~\ref{nondNb}-d), 
DBX crossover cannot be applied because
no individual actually dominates anyone in the population!
The actual rate of application of the DBX operator 
is maximal about the moment when the half of the
population gets the rank~1 (Fig.~\ref{nondNb}-c).

One possible improvement would be to increase the rate of
application of DBX operators at the beginning of evolution 
by mating each non-dominated individual with not only one but several 
individuals it dominates.

Note that dynamic behavior of the populations described here above 
and, in particular,
the disappearance of the dominated individuals is due to the 
the replacement procedure in NSGA-II, and might not be so
prominent with other EMAs.
It is hence worth investigating the use of DBX restricted mating
with other EMAs paradigms, such as SPEA2 or PESA, for instance.

Furthermore, there exist other crossover operators used in EAs (such
as SBX~\cite{book}, for example) that
could be applied together with dominance based restricted mating, 
instead of BLX-$\alpha$.

One more issue, that needs to be thoroughly investigated, is the
efficiency of the DBX strategy when solving problems with more than
two objectives. Indeed, in such situations, the
``visual'' analysis of the populations dynamics performed in the present work 
will not be possible anymore, and our plans include the use of so-called 
{\it running metrics}~\cite{DebJain:2002}.

Indeed, the present study could use
some performance metrics as well, instead of the graphical plots of
the population dynamics in the objective space. 
However, in our opinion, such presentation is much clearer and makes
things more explicit, hence it is more appropriate to the introduction
of the new operator. 

But probably  the most important issue is to find a way to actually
benefit from the reported acceleration in practice. Whereas its
usefulness is obvious for costly applications, where the number of
evaluations has to be limited, the question remains of how to detect
that the algorithm is getting stuck, thus saving the computational
time that restricted mating allows us to spare.
An efficient stopping criterion is needed before any actual benefit
can be gained from  the observed acceleration.  
Such criterion has been proposed by the first author
\cite{these-Olga,MOPGP}, but it was beyond the scope of this paper. 
Nevertheless, together with the DBX operators described in this paper, it
allowed an actual saving  of about 8\% of the computation time. 
However, further evaluation and refinement of that stopping criterion
are still needed before definite strong conclusions can be drawn.


\end{document}